\documentclass[journal]{journal}

\usepackage{graphics} 
\usepackage{epsfig} 
\usepackage{color,hyperref}
\usepackage{amsfonts}
\usepackage{fancybox}
\usepackage{lipsum}
\usepackage{graphicx}
\usepackage{amsmath} 
\usepackage{subfig}
\usepackage[font=small]{caption}

\usepackage[linesnumbered,ruled]{algorithm2e}

\begin{document}	
\title{\LARGE \bf
Joint Path and Push Planning Among Movable Obstacles
}


\author{Victor Emeli$^{1}$ and Akansel Cosgun$^{2}$
        \thanks{$^{1}$Georgia Institute of Technology,
         USA.} \thanks{$^{2}$Monash University, Australia.}}

\maketitle
\thispagestyle{empty}
\pagestyle{empty}

\begin{abstract}
\textbf{This paper explores the Navigation Among Movable Obstacles (NAMO) problem and proposes joint path and push planning: which path to take and in what direction the obstacles should be pushed at, given a start and goal position. We present a planning algorithm for selecting a path and the obstacles to be pushed, where a Rapidly-exploring Random Tree (RRT)-based heuristic is employed to calculate a minimal collision path. When it is necessary to apply a pushing force to slide an obstacle out of the way, the planners leverage means-end analysis through a dynamic physics simulation to determine the sequence of linear pushes to clear the necessary space. Simulation experiments show that our approach finds solutions in higher clutter percentages (up to 49\%) compared to the straight-line push planner (37\%) and RRT without pushing (18\%).}
\end{abstract}
\section{Introduction}

Robots would be much more useful if they had the capability to move obstacles out of the way. Such robots can be deployed in search and rescue scenarios, where they reach to places unreachable by humans and assist in the removal of rubble from disaster areas. In an attempt towards this vision, we explore the Navigation Among Movable Obstacles (NAMO) problem in which the robot attempts to navigate from one side of an environment towards reach a goal position in a reconfigurable environment, while manipulating obstacles along the way. We assume a 2D environment where the agent is considered to have a polygon-shaped footprint and can push any object in the environment.

In this paper, we propose a planning algorithm which is a minimal collision path planner using a RRT-based heuristic. Our approach attempts finding a feasible path by making the area of the agent's footprint iteratively smaller, while at the same time planning for pushes that would clear the space for the actual size of the agent. In simulation experiments, we benchmark our approach against the straight-line navigation with pushes as well as a standard collision-free RRT. By interleaving path and push planning, we reduce the algoritmic complexity of the problem is reduced greatly.

The organization of this paper is as follows. We examine the relevant literature in Section~\ref{sec:related-work}, including non-prehensile manipulation, and physics simulators. Section~\ref{sec:problem_description} defines the problem domain and Section~\ref{sec:proposed-algorithm} describes our proposed algorithms in detail. Section~\ref{sec:experiments} discusses experimental results produced in simulation with variations in domain complexity. Section~\ref{sec:conclusion} places our planners in the broader context of clutter manipulation.

\begin{figure}[t!]
	\centering
	\includegraphics[clip, trim=0cm 0.2cm 0cm 0cm, width=0.49\textwidth]{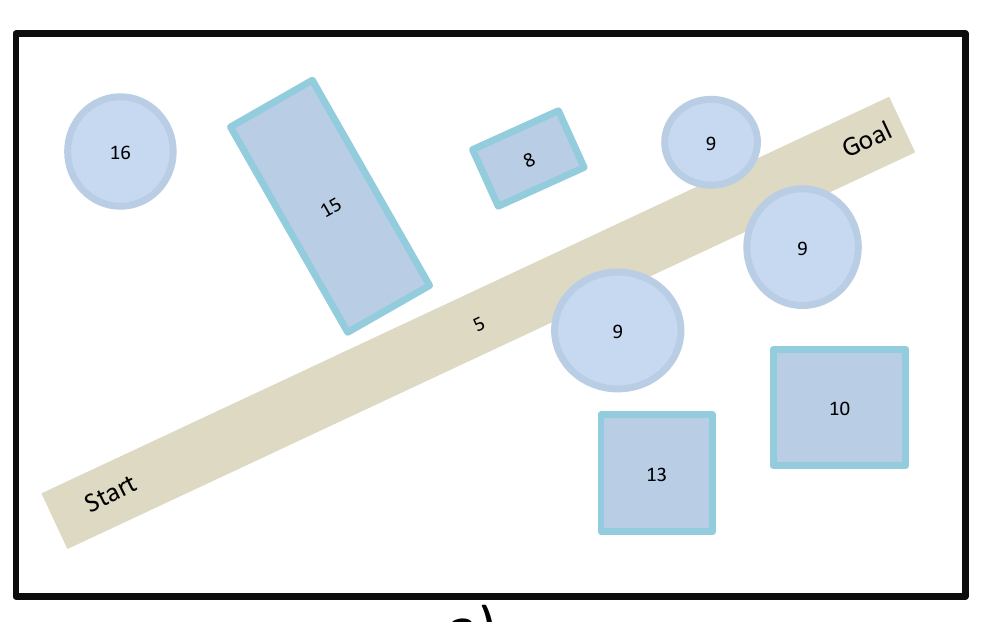} 
	\includegraphics[clip, trim=0cm 0.2cm 0cm 0cm, width=0.49\textwidth]{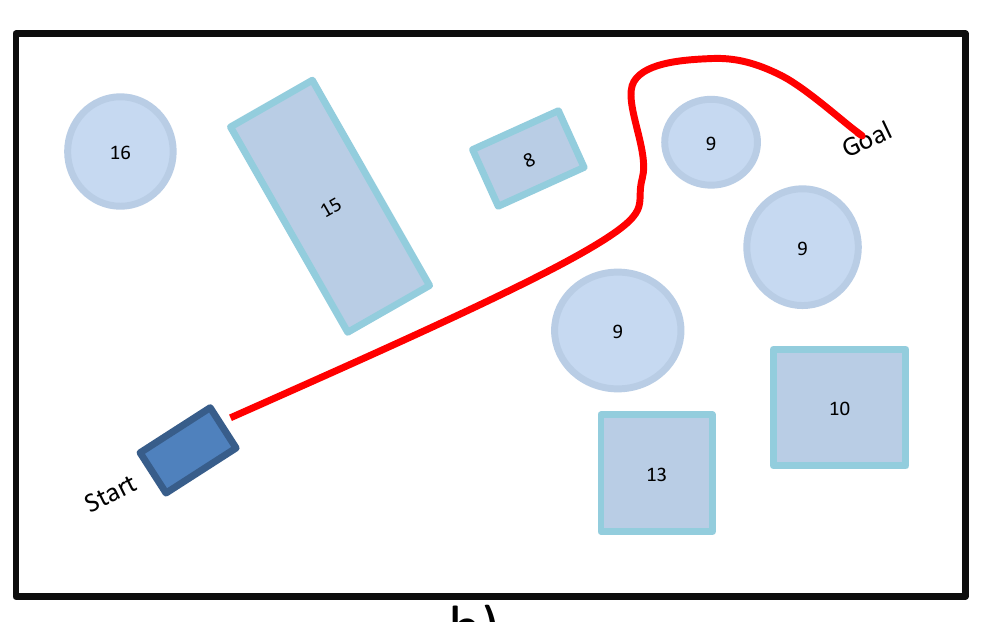} 
	\caption{We study the problem of navigation among movable objects. We model the environment as a 2D world with polygonal objects and propose two algorithms for pushing obstacles to make enough space for the robot: 1) straight line navigation with push planning (Top) and 2) RRT-based iterative minimal collision with nominal pushing using a reduced area rigid body (Bottom).}
    \label{fig:intro}
\end{figure}

\section{Related Work}
\label{sec:related-work}

Our problem definition falls into the arena of manipulation planning.  This involves planning the motion of a robot and the manipulation of one or more objects in the presence of clutter.  This area includes manipulation planning among movable obstacles (MAMO)[6][16], rearrangement planning (RP)[11, 12], and navigation among movable obstacles (NAMO)[13,14]. Wilfong et al. [7] showed that NAMO is NP-hard, and that rearrangement planning is PSPACE-hard. The complexity arises from the high dimensional search space and the constraint that objects only move as a consequence of the robot’s actions. Alami et al. [8] classifies robot actions into two categories. Transit actions, which are collision-free robot motions, and transfer actions, which manipulate objects. Planning transit actions is a classical robot motion planning problem, and planning transfer actions requires additional intelligence about the mechanics of manipulation.

Non-prehensile pushing allows for more flexibility and efficiency in completing a task~\cite{cosgun2020,brock,choi}.  Pushing is desirable because it is quicker to execute, may reduce uncertainty~\cite{dogar2010} and be exerted on heavy or multiple objects at the same time.  In rearrangement planning it is more useful to push multiple objects at the same time than sequentially.  Ben-Shahar et al. [9] illustrates a rearrangement planner that allows the concurrent pushing of multiple objects. The algorithm performs a hill-climbing search on a sampled-based representation of the configuration space. It minimizes a cost function that represents the minimal cost to reach the goal. This cost is computed offline on the discrete search space using a reverse pushing model that compensates for multi-object contacts and non-quasistatic physics. Computing such a general reverse pushing model, however, is difficult and only a simplified model is presented.    
Our approach presents a high level planner that attempts to enable an agent to take a straight line path from a start to goal configuration in a cluttered environment by pushing movable obstacles out of the way while transporting a large object.  This capability is beneficial in situations when a collision-free path is either too expensive to travel or does not exist at all.  The planner attempts to accomplish this by (1) selecting a straight line path from the start to goal position by leveraging a heuristic that provides a path with minimal obstacle overlap in the environment and (2) determines a sequence of pushing actions that creates a straight unobstructed path for the agent to traverse from start to goal.  The utilization of a straight line path, as opposed to trajectories, greatly reduces the search complexity of this algorithm.  Despite significant differences, some of the concepts in this work are closely related to [15] which applies means-end analysis in order to efficiently compute plans for displacing objects with multiple interactions. We present a similar process of reverse search in our domain.  This contribution builds on our work in [1] and [4] where nonprehensile push planning with dynamic obstacle interaction was leveraged to place large objects on a cluttered table.  

\begin{figure*}[h]
\subfloat[]{\includegraphics[clip, trim=0.3cm 0.3cm 17cm 0.3cm, width=0.3\textwidth]{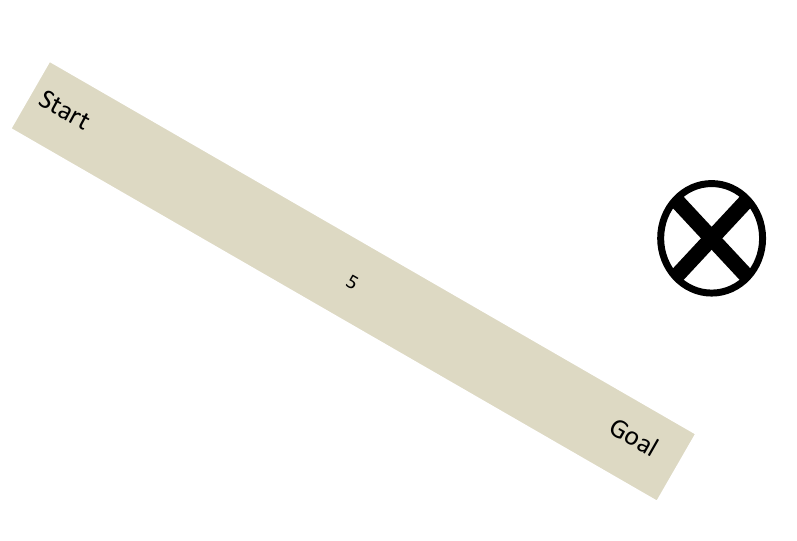}}
\subfloat[]{\includegraphics[clip, trim=8.25cm 0.2cm 9.2cm 0.1cm, width=0.3\textwidth]{./Images/fig2.pdf}}
\subfloat[]{\includegraphics[clip, trim=16.4cm 0.2cm 0.3cm 0.1cm, width=0.33\textwidth]{./Images/fig2.pdf}}
\caption{a) Path footprint b) Current Environment c) Heuristic path placement after convolution with the environment.}
\end{figure*}


This work is also related to combining task and motion planning (TMP), which is still an open problem in the research community and is key to implementing successful mobile manipulation solutions [21-25]. The goal of an agent moving an object from a start to a goal position in a cluttered room is a high level discrete task, which is symbolically represented in TMP.  The actual motion planning that accomplishes this effort takes many factors into consideration, which include continuous reasoning, uncertainty, and replanning.  This is the algorithmic portion of TMP.  The motion planning and physics-based object interaction is accomplished in simulation in this work.  Next steps will encompass implementation on a physical robot, which will include lower-level planning to facilitate the human-robot physical interactions and the necessary object manipulations in the real world.  This contribution builds on our work in [1] and [4] where nonprehensile push planning with dynamic obstacle interaction was leveraged to place objects on a cluttered table.

\section{Problem Description}
\label{sec:problem_description}
Given an example of a rectangular room, $R$, defined by bounding corners $((x_{\min},\ y_{\min}), (x_{\max}, y_{\max}))$ and an assortment of object shapes $O = \{0_{1},\ 0_{2},\ .\ .\ .\ ,\ 0_{n}\}$. The first $n-1$ shapes describe the objects residing in the room and $0_{n}$ defines the shape of a virtual path, which represents a straight line path from the starting point to the goal. Let $q= (P_{1},\ P2,\ .\ .\ .\ ,\ P_{n})$ define the varying poses of all the objects, where $P_{j} = \{(x,\ y)\ :\ x\ \in \mathbb{R},\ y\ \in \mathbb{R},\ (x,\ y)\ \in\ 0_{j}\}$ is the set of all points occupied by object shape $0_{j}$. A linear pushing action is defined as $u_{j,k} = (0_{j},\ \phi_{k},\ d_{j,k})$), where the action is exerted on object $0_{j}$ in the direction $\phi_{k}$ for a distance $d_{j,k} =d(0_{j},\ \phi_{k},\ q) >0$ with a constant velocity $\epsilon>0$. The dynamic interactions between objects is governed by a function $f$, determined by the physics simulator, where $\mathrm{q}=f(O,\ q,\ u)$ .

For the straight line planner, finding a straight line path from a starting point to a goal in a cluttered room requires an agent to find a sequence of push actions $U = (u^{1},\ u2,\ .\ .\ .\ ,\ u^{t})$ that will result in a state $q_{\mathrm{e}nd}$,  

For the minimal collision planner, finding a minimal collision path from a starting point to a goal in a cluttered room requires the RRT~\cite{rrt} algorithm to calculate discrete waypoints in the SE(2) state space and an agent to find a sequence of push actions $U = (u^{1},\ u2,\ .\ .\ .\ ,\ u^{t})$ that will result in a state $q_{\mathrm{e}nd}$,where three conditions are satisfied:

\begin{enumerate}
\item No objects intersect: $\bigcup_{P_{i},P_{j}\in q_{\mathrm{e}nd},i\neq j}(P_{i}\cap P_{j})=\emptyset$
\item Objects are within the room wall borders: $x_{\min} < x < x_{\max}, y_{\min}<y<y_{\max}\forall(x,\ y)$ in $P\in q_{\mathrm{e}nd}.$
\item Objects are stationary: $\dot{q}=0.$ 
\end{enumerate}

The potential search domain is infinite if there are no action constraints. The domain is limited to allow only one push per object $U$, such that $0_{i}\neq 0_{j}$ for every pair $(u_{i,k},\ u_{j,l}) = ((0_{i},\ \phi_{k},\ d_{i,k}),\ (0_{j},\ \phi_{l},\ d_{j,l}))$ in $U.$ Even in this domain, an exhaustive search is inefficient. Given a set of $n$ objects and $g$ pushing angles, the branching factor would be {\it ng} and total nodes in the tree would be $(ng)^{n}$. For $n=g=10$, an exhaustive search tree produces $10^{20}$ nodes, a space too significant for standard hardware. This formulation creates a maximum of $n$ pushes in any completed plan. We allow $k$ pushes per object, where $k$ is the number of objects initially overlapping the start to goal path, resulting in a maximum number of {\it kn}. By combining a reduced action space with informed heuristics and minimal collision path, we attempt to simplify the solution. In order to achieve generality and efficiency, we separate the task of moving an object in a cluttered environment into two stages: The first stage determines the pose/shape $P_{n}$ for the proposed path $0_{n}$ that the human-robot team will attempt to traverse. The second stage finds the set of push motions $U$, that satisfy the three requirements listed above. In order to discover shorter plans, we iterate stages 1 and 2 gradually towards more complex plans.

The start and goal position are sampled with the bounds of the rectangle (the room).  The samples are allowed anywhere on the y-axis, however only within the first 2 cm (start positions) and the last 2 cm of the x-axis (goal positions).  Also the sample is only valid if the proposed straight-line path (for the straight line planner) or the rigid body (for the minimal collision planner) fits within the rectangle.

\begin{figure*}[h]
	\centering\includegraphics[clip, trim=0cm 0.25cm 0cm 0cm, width=.9\textwidth]{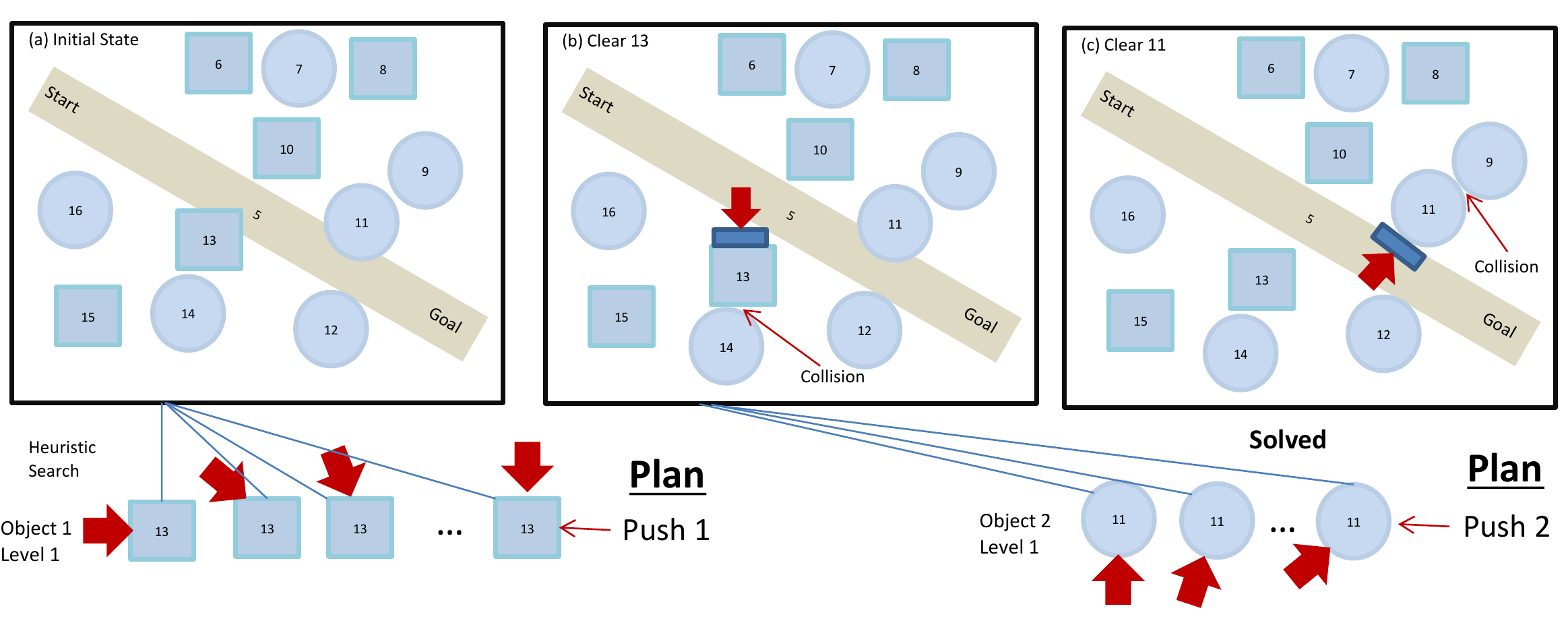}
	\caption{An illustrative example that demonstrates the push planning algorithm for a Level 1 search.}
\end{figure*}

\section{Proposed Algorithm}
\label{sec:proposed-algorithm}
\subsection{Goal Configuration Calculation}

The technical details of the goal configuration calculation for the straight-line push planner is covered in our previous work [1]. This approach uses a heuristic that attempts to sample object placement locations with the least amount of overlap with clutter using a convolutional technique.  In our work, we modify the algorithm to sample straight-line start to goal configuration paths with the least amount of overlap with obstacles for the straight-line planner.  Figures 2(a), (b), and (c) shows an example of a sampled path footprint, a cluttered environment, and the heuristically determined path footprint pose in the cluttered environment.  Given a pose sample, we present a solution for push planning that clears the path footprint in Sections IV-B and IV-C.  
For the minimal collision planner, we utilize The Open Motion Planning Library~\cite{ompl} (OMPL) to attempt to calculate a collision-free path from a start to goal configuration in clutter using the RRT-Connect~\cite{rrt-connect} algorithm and a reduced sized rigid body.  RRT-Connect is a probabilistically complete randomized algorithm used to solve single-query path planning problems. This algorithm incrementally builds two RRTs rooted at the start and the goal configurations. The human-robot team is represented as a rigid body with area $T$.  Initially, the RRT-Connect algorithm attempts to calculate a collision-free path from the start to goal configuration.  If it is successful, then the algorithm succeeds.  If not, it reduces the area of the rigid body by 10$\%$, where $A_{n}$ = $0.9A_{n-1}$, and the algorithm is executed again.  This is repeated recursively until a valid path is achieved or until the size is reduced to a point robot and no solutions were found.  Samples of the discrete start to goal waypoints of the rigid body are inputted as the goal configuration in the push planner.  The rigid body's discrete poses moves through the configuration space at each sampled point.  Since a version of the rigid body with a smaller area is able to navigate through a collision-free path, the larger rigid body's path would have minimal overlap with clutter in the environment.

\begin{figure*}[h]
	\centering\includegraphics[width=.99\textwidth]{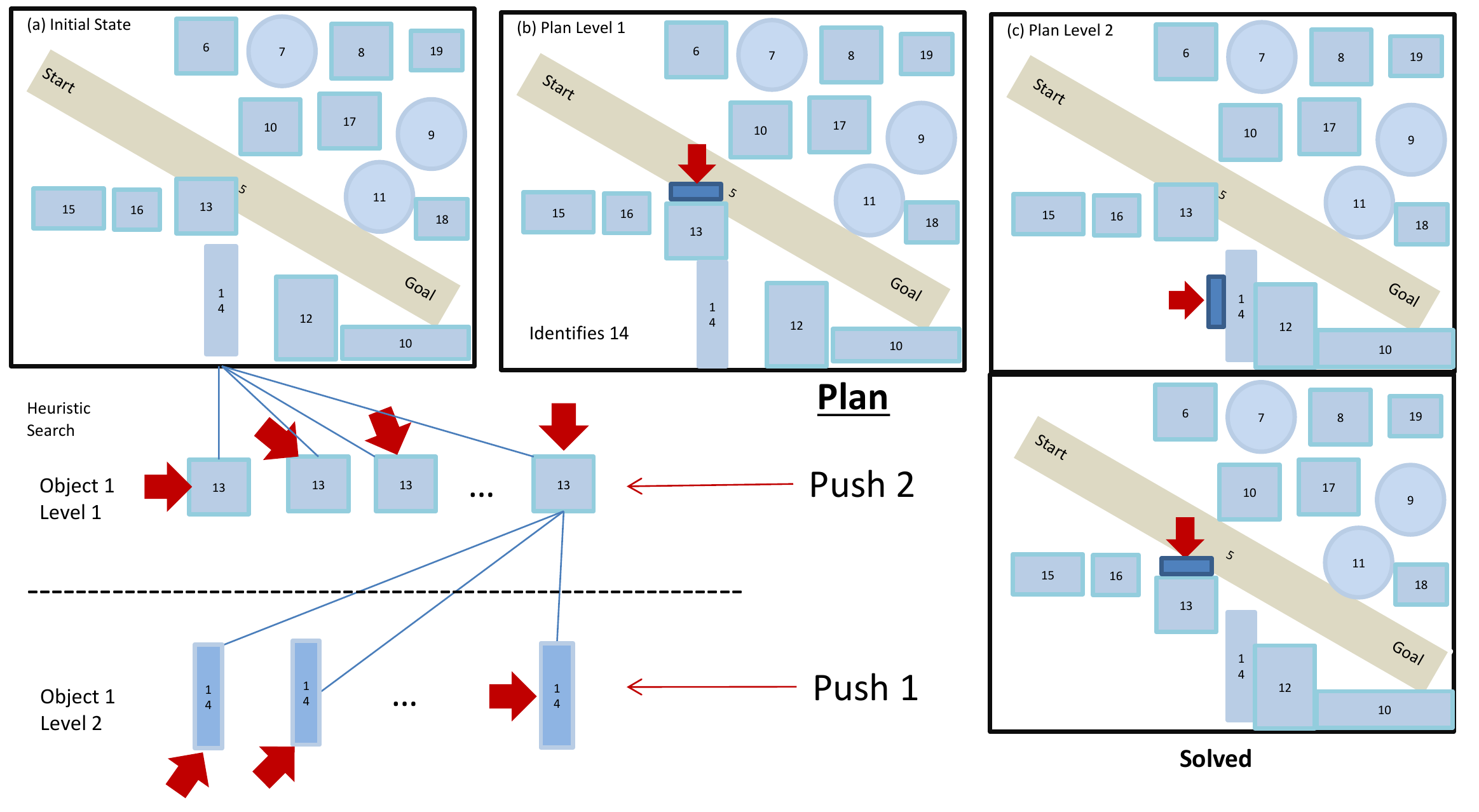}
	\caption{An illustrative example that demonstrates the push planning algorithm for a Level 2 search.}
\end{figure*}

\subsection{Path Clearing Planner}
The straight-line planner and minimal collision planner both utilize the same push planning algorithm.  The following example applies to the straight-line planner, however the same principles apply to the minimal collision planner.  

Given a potential straight line start to goal path pose $\hat{P}_{n}$ we introduce a planning algorithm that clears the path defined by $(\hat{P}_{n},\ 0_{n})$ . First, we present the algorithm for the case where two objects {\it o}13 and {\it o}11 overlaps with the path $0_{n}$.  Figure $3$ illustrates the search tree for this example.  This case only requires one level (Level 1) of push planning for each object. Next, we expand this approach to one object {\it o}11 requiring two levels (Level 2) of push planning.  The details of this algorithm are presented in Section IV-C and Figure $4$ illustrates the search tree for this example. The planner uses Breadth First Search (BFS) to attempt the possible pushing actions $U_{o_{1}} = \{u_{1,k} = (0_{1},\ \phi_{k},\ d_{1,k})\}$. It determines the pushing distance $d_{1,k}$ through the push termination conditions: (1) Any object colliding with the room wall or making contact with $ 0\in O$ that cannot be pushed. (2) $ 0_{1}\cap 0_{n}=\emptyset$. The interactions during the motion returns the resulting $q_{n\mathrm{e}w}$ and the first blocking object $0_{j}$ to the planner. In order to constrain the large set of possible pushes, our algorithm begins by investigating only the set of pushing actions $U_{o_{1}}$ on overlapping  object {\it o}13. Let's suppose every push $u_{1,k}$ terminates following rule (1). After each push of {\it o} 13, we detect any object $0_{j}$ that blocks the motion of {\it o}13.  In this example, we are able to clear {\it o}13 from the path during the first level of push planning.  The algorithm is repeated for {\it o}11 that also overlaps the path and it is able to also clear the path during the first level of push planning.  Next we consider the case of one overlapping object {\it o}13 that requires two levels of push planning to clear the path.  Following each push of {\it o}13, we detect any object $0_{j}$ that blocks the motion of {\it o}13.  After searching all single pushes of {\it o}13, the planner resets to the initial configuration $q_{init}$ and searches over the pushing actions $U_{o_{j}}$ of each blocking object $o_{j}$. Given the subsequent configuration from each push of $0_{j}$ the planner backtracks to {\it o}11 and searches over the pushes $U_{o_{1}}$, testing if any clears {\it o}13 from the path footprint. This approach guides the search by means-end analysis. The planner attempts to remove the blocking object $0_{j}$ from the path of {\it o}13. Since additional objects may constrain the movement of object $0_{j}$ the planner recursively follows this procedure until it clears the path footprint or reaches a maximum depth $L_{MAX}$.  In this case, {\it o}14 is cleared from the path of {\it o}13.  This results in a two level, two push plan to clear the path. In the following section we extend this algorithm to multiple objects overlapping the straight-line path or minimal collision path.

\subsection{Iterative-Deepening Breadth-First Search}

In the case of multiple overlapping objects, the push planner proceeds as follows. A path footprint is selected by rejection sampling from the probability distribution in Sec. IV-A. Each object that overlaps the path is a {\it sub}- {\it goal}. For each sub-goal, the planner applies Algorithm 1 in [1]. If a goal test succeeds then the planner continues with the next overlapping object starting from $q_{init}'$, where the sub-goal is satisfied. Algorithm 2 in [1] shows how the planner attempts prior pushing actions to evaluate if a sub-goal has been reached. A sub-goal succeeds if the number of overlapping objects has been reduced. Each configuration likely has many solutions. Our planner uses a heuristic to guide it towards solutions with fewer pushes. We extend our algorithm to use Iterative-Deepening Breadth- First Search (IDBFS). We add an outer loop around Algorithm 1 which iteratively increases the maximum allowed tree depth from $0$ to $L_{\max}$. IDBFS allows our BFS planner to run on a number of different straight line start to goal paths in an attempt to find shorter plans which will require fewer pushes. Algorithm 3 in [1] explains the full IDBFS Push Planner, which takes into account multiple objects overlapping the path.  Iterative-Deepening Depth-First Search (IDDFS)~\cite{iddfs} inspires our search, which maintains the minimal solution length of BFS while taking advantage of Depth-First-Search's efficiency.



\section{Experiments}
\label{sec:experiments}

Our planners use OMPL to calculate collision-free or minimal collision paths from a start to goal configuration and the open source $2\mathrm{D}$ physics engine $\mathrm{B}\mathrm{o}\mathrm{x}2\mathrm{D}$~\cite{box2d}. OMPL implements sampling-based motion planning including probabilistic roadmaps and tree-based planners. $\mathrm{B}\mathrm{o}\mathrm{x}2\mathrm{D}$ models contact, friction and restitution as well as collision detection. We modeled room objects as rigid bodies of convex polygonal objects with equal densities. Push actions were applied with a rectangular rigid body $0_{g}$ of dimension $1\mathrm{c}\mathrm{m}\times 4\mathrm{c}\mathrm{m}$. To apply $u_{i,k}=(0_{i},\ \phi_{k},\ d_{i,k})$ , first $0_{g}$ is placed centered at the center of mass of $0_{i}$ with orientation $\phi_{k}$ and gradually moved along $(\phi_{k}\ -\pi)$ direction while checking collision between $0_{i}$ and $0_{g}$. At the configuration where $P_{i}\cap P_{g} = \emptyset$, a final test between $0_{g}$ and all other objects in $O$ verifies whether the rigid body can be placed without collision. If the configuration is collision free, the pushing action is feasible and $0_{g}$ is moved in $\phi_{k}$ direction at a constant velocity.

The algorithm is tested using gradually increasing clutter percentages of objects. Clutter percentage is defined as the ratio of the area occupied by the objects to the total room area. The virtual room size was $38\mathrm{c}\mathrm{m}\times 19\mathrm{c}\mathrm{m}$ for all experiments, the object shapes are 20 squares that are near evenly distributed throughout the room.  The length and width of each square is increased by $.25\mathrm{c}\mathrm{m}$ respectively for each clutter percentage increase.  


\begin{figure}[ht!]
\subfloat[]{\includegraphics[clip, trim=0.3cm 0.3cm 16.45cm 0.3cm, width=0.5\textwidth]{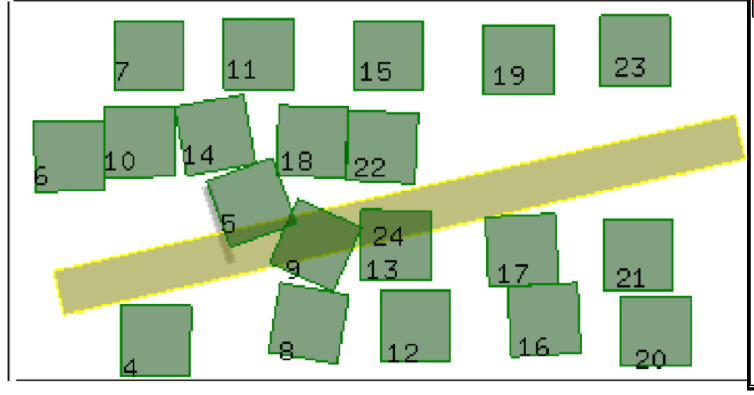}} \\
\subfloat[]{\includegraphics[clip, trim=7.8cm 0.2cm 8.45cm 0.2cm, width=0.5\textwidth]{./Images/three_images.pdf}} \\
\subfloat[]{\includegraphics[clip, trim=15.9cm 0.32cm 0.3cm 0.25cm, width=0.5\textwidth]{./Images/three_images.pdf}}
\caption{a) Straight-line push planner attempts to clear a path, but fails at 43\% room clutter. b) Minimal collision planner utilizes RRT Connect to produce a valid path heuristic using a reduced area rigid body at 43\% room clutter. c) Minimal collision planner uses the heuristic to execute a valid plan using the footprint of the robotic agent.}
\end{figure}

The path for the straight-line planner has a width of $1.5\mathrm{c}\mathrm{m}$ and a length of $17.5\mathrm{c}\mathrm{m}$. The simulator provides the room configuration $q_{init}$ as input to the planning algorithm for path placement. If a path placement candidate is found, the simulator executed the pushes and attempts to clear a straight line path before selecting a new path. If the algorithm failed, the simulator reset the room to the initial configuration and repeated the procedure.

The path for the minimal collision planner is provided as discrete poses by the RRT planner after utilizing a reduced area version of the original rigid body.  The original rigid body has a length of $4.5\mathrm{c}\mathrm{m}$ and a width of $2.25\mathrm{c}\mathrm{m}$.  It utilizes the poses provided by the RRT to attempt to traverse from the start position to the goal position.

To make the algorithm practical for real world implementation, the maximum allowed tree level was set to 3, the maximum number of path configuration samples per allowed tree level was 20 and the push angle resolution was $\pi/12$, resulting in 24 push directions. If a plan was not found for all the candidate positions, the algorithm terminated with no solution. We performed 10 trial runs for each clutter percentage. Figure 5 illustrates a room configuration with 43\% clutter.  a) The straight-line planner was unable to find a solution at this clutter percentage.  However, after reducing the original rigid body area by 84\% and running the RRT (b), the minimal collision planner was able to find and execute a valid plan (c).  Table~\ref{lab:table} shows the results for the collision-free planner (RRT Connect), the straight-line push planner, and the minimal collision push planner at clutter percentages ranging from 18\% to 56\%.


\setlength{\tabcolsep}{0.15cm}
\begin{table}[ht!]
\centering
\begin{tabular}{|c|c|c|c|c|c|c|}
\hline
{\small Clutter}     & {\small RRT-Connect} & {\small Straight-Line} & {\small Min. Collision}\\ \hline
18\% & 10 & 10 & 10 \\ \hline
37\% &0 & 4 & 10 \\ \hline
43\% &0 & 0 & 7  \\ \hline
49\% &0 & 0 & 5 \\ \hline
56\% &0 & 0 & 0 \\ \hline
\end{tabular}
\caption{Experimental results shows the number of successful plans out of 10 trials vs. clutter percentages for each planning algorithm (RRT-Connect, straight-line, and minimal collision.}
\label{lab:table}
\end{table}

The RRT planner's maximum clutter percentage for finding a valid plan for the original rigid body is 18\%. The straight-line push planner's maximum clutter percentage is double that of the RRT planner at 37\%.  It was not able to find a solution at the next clutter percentage of 43\%.  This can be attributed to the fact that the start to goal path encompasses a long rectangular path which is almost equal to the total length of the room.  This greatly increases the number of potential overlapping objects and decreases the chance of clearing all of them from the straight-line path placement candidate.  Since the length of the room is greater than the width of the room and the path's length is almost equal to the room's length, the number of possible unique poses are also constrained.  The straight line path constraint simplifies the search space at the cost of reducing the ability of the planner to find a solution.  The minimum collision push planner is able to find a valid path up to a clutter percentage of 49\%.  The success at this higher clutter percentage can be attributed to the fact that the planner does not have to take a straight-line path, which may be sub-optimal at times.  The RRT-based heuristic guides the push planner through a non-linear path that has minimal collisions, which is an advantage over the straight-line planner.   

\section{Conclusion}
\label{sec:conclusion}

We have presented an algorithm for clearing objects in a room to facilitate navigation among movable obstacles. Exploiting the candidate path pose generating heuristics assist in finding shorter solutions by providing a variety of path candidates to the planner. Applications to a physical robot will benefit from this bias towards simpler solutions. Plans with a greater number of push actions during real world operation induce a greater chance of divergence from the simulated physics used by the planner. Constraining the number and distance of these actions  reduces this error and agent can perform the task more robustly.  The results of the experiment suggest that in practice the agent can first attempt to find a collision-free path using a standard RRT. If this is not feasible, then it can then execute the straight-line push planner, since moving in a straight line while pushing objects out of the way will expend the least amount of energy. If the straight-line push planner fails, then the RRT-based minimal collision push planner can be employed because it has demonstrated success at the highest clutter percentage, but would result in the most energy consumed during execution of the plan.

\end{document}